\documentclass[conference]{IEEEtran}
\IEEEoverridecommandlockouts

\usepackage{cite}
\usepackage{amsmath,amssymb,amsfonts}
\usepackage{algorithmic}
\usepackage{graphicx}
\usepackage{url}
\usepackage{textcomp}
\usepackage{xcolor}
\def\BibTeX{{\rm B\kern-.05em{\sc i\kern-.025em b}\kern-.08em
    T\kern-.1667em\lower.7ex\hbox{E}\kern-.125emX}}
\begin{document}

\title{
On-board ML for Trace Gas detection in Imaging Spectroscopy data
\thanks{This work was carried out at the Jet Propulsion Laboratory, California Institute of Technology, under a contract with the National Aeronautics and Space Administration (80NM0018D0004). 
}
}

\author{\IEEEauthorblockN{Vít Růžička\IEEEauthorrefmark{1},
Adam Chlus, Andrew Thorpe and
David R. Thompson}
\IEEEauthorblockA{Jet Propulsion Laboratory,
California Institute of Technology,  4800 Oak Grove Drive, Pasadena CA, USA\\
Email: \IEEEauthorrefmark{1}ruzicka@jpl.nasa.gov}}
\maketitle

\begin{abstract}
Data collected during aerial and spaceborne imaging spectroscopy campaigns enables the detection of transient events such as trace gas emissions. However, current processing pipelines depend on slow, on-the-ground processing, which delays the time to information of each detected event and prohibits immediate follow-up actions. During the Tokyo Field Campaign of March 2026, we explored on-board processing of Imaging Spectroscopy data from the equipped AVIRIS-5 sensor. Due to communication bottlenecks, full datacubes cannot be downlinked immediately during the flight. Instead we downlink the potential events predicted by our efficient and small machine learning model. We show the first on-board detection of methane point source emission with Imaging Spectroscopy data using Edge ML.
\end{abstract}

\begin{IEEEkeywords}
machine learning, on-board, imaging spectroscopy, trace gas detection.
\end{IEEEkeywords}


\section{Introduction}
Imaging spectroscopy (also known as Hyperspectral) sensors generate large quantities of remote sensing data, which is difficult to process in real time on-board of the spaceborne and aerial platforms. Instead, the data is downlinked to the on-the-ground stations during flyover, or at the end of the flight day during an aerial campaign. Work of \cite{thompson2015real_time_mf_on_plane} has shown that real-time computed matched filter (MF) products of trace gases can be used as an effective guidance for the flight operators and that the overall number of detected events can be increased by having these live preview products. However, this depends on manual inspection by the flight crew. Recent work of \cite{ruzicka2026trace_gases} has shown that machine learning (ML) models can be used in combination with these enhancement products to reliably detect point source events of CH$_{4}$, NH$_{3}$, NO$_{2}$ and CO.

In this paper, we describe the successful deployment of ML models for the on-board detection of trace gas events during an aerial imaging spectroscopy campaign. The recent Tokyo-FC campaign\footnote{Tokyo-FC Project Plan in: \url{https://espo.nasa.gov/sites/default/files/documents/2025-12/Tokyo-FCProjectPlan.pdf}} collected data with the AVIRIS-5 sensor over Japan between 9-31 March of 2026. We used a small and efficient ML model to process captured data on-board during the flight. Model predictions were vectorised and the data in each vector was extracted and downlinked through a constrained communication channel enabling real-time inspection of the results. Prior to the flight, we tested our machine learning model with a simulated data processing pipeline on a testbed device consisting of Jetson Orin Nano Super board.

\begin{figure}[!t]
    \centering
    \includegraphics[width=0.95\linewidth]{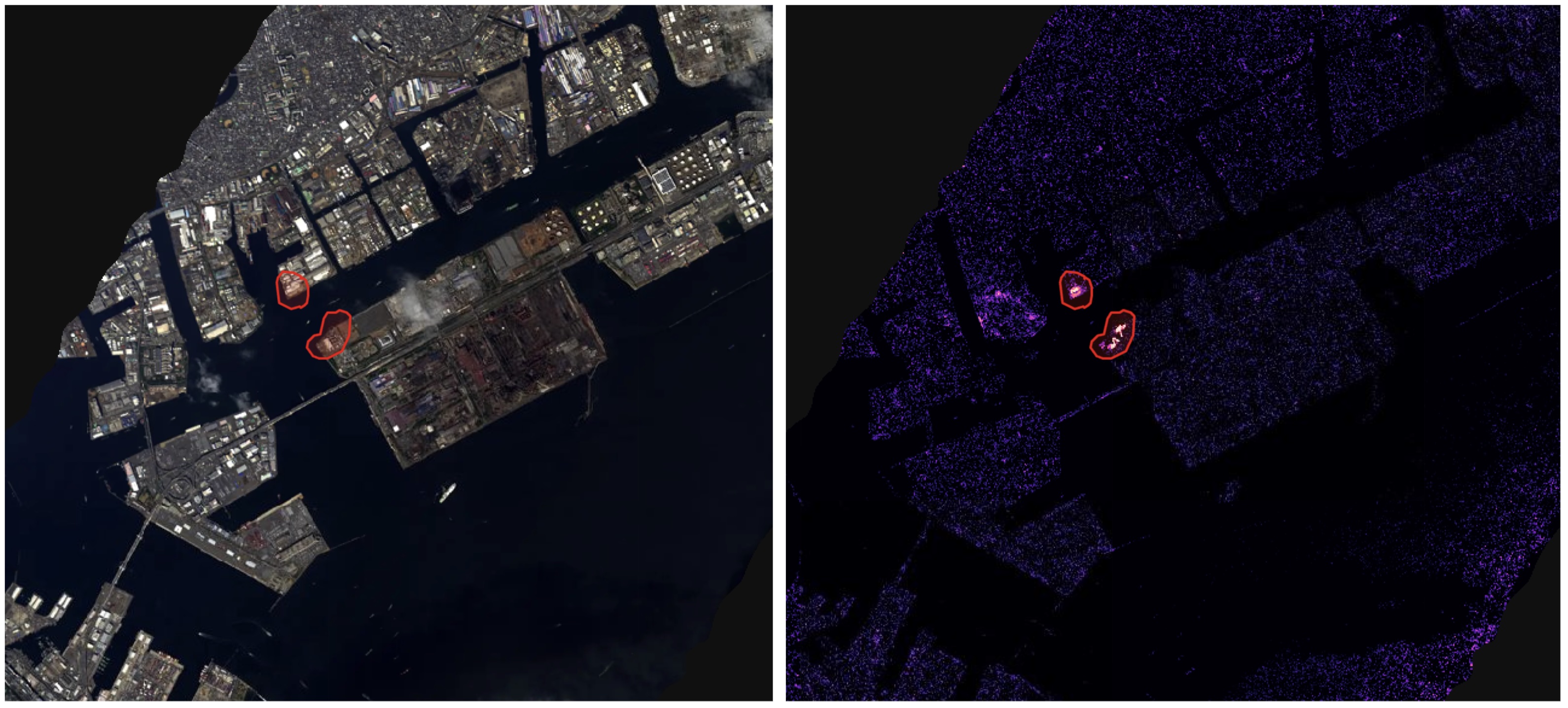}
    \caption{Methane point source event detected on-board during the Tokyo-FC campaign. We show RGB bands (left) and on-board computed methane enhancement product (right), red outline shows our model prediction.}
    \label{fig:detection}
\vspace{-3mm}
\end{figure}

Finally, in Figure \ref{fig:detection} we report successfully detected CH$_{4}$ point source event observed on the 11th March flight over a power plant in Kawasaki. To our knowledge, this is the first instance of detecting trace gas emission on-board of a real mission using ML models. This is a significant step, which promotes instrument autonomy and may enable future fully autonomous processing of data in low compute and communication bottleneck scenarios.

\vspace{-0.7mm}
\subsection{Related Work}

Imaging Spectroscopy data consists of many spectrally narrow bands, sensors considered in this work span between 380 to 2500nm. As such, they offer visibility of a number of trace gases, nonetheless robust detection of these events remains challenging. 
The work of \cite{HyperspectralViTs} first proposes the approach of using ML models for on-board detection of methane leak events. It defines two ways of processing data, either using end-to-end ML models which process radiance data directly, or via dependence on MF products. The work of \cite{herec2025optimizing} follows up in the direction of speeding up MF computations and then using ML models to reduce the number of false alarms. 

\vspace{-0.2mm}

\section{Methodology}

\subsection{Trace gases detection in Imaging Spectroscopy data}

The first step in processing Imaging Spectroscopy data on-board is the conversion from digital number (DN) to radiance values. We follow the publicly released data processing pipeline of the EMIT mission as a proxy for our initial on-the-ground experiments. The process is in greater detail explained in \cite{thompson2024orbit_EMIT} and the code repository of\footnote{\url{https://github.com/emit-sds/emit-sds-l1b}} - raw data is called L1A and radiances are the L1B product.

\subsection{Proxy environment}

We build the testbed environment using the NVIDIA® Jetson Orin™ Nano Super board (in 30W mode). It has a 1.7 GHz 6-core Arm Cortex CPU, Ampere arch. 1020 MHz GPU and 8GB RAM.

We start with the full EMIT on-the-ground pipeline described in \cite{thompson2024orbit_EMIT}. First, we skip several steps in order to speed up the processing, while observing minimal differences in the computed products. In all cases, we avoid the order sorting filter (OSF) correction, as it affects bands outside of typical methane visibility (2100-2500nm). We also explore skipping bad pixel and ghost correction. 
We vectorise computations and merge repeated operations where possible. 
We highlight that the full on-the-ground pipeline strives for accurate science ready data products, meanwhile for fast on-board detection of very large methane leak events, some shortcuts can be taken.
As a second optional step, we compute the matched filter product. We use a non-iterative spectrally wide variant. 
Finally, in the third step, we use a machine learning model either on the radiance data, or on the matched filter product computed for the same location. For end-to-end radiance data processing, we leverage models published in \cite{HyperspectralViTs}, namely the HyperEfficientViT (ConvUp) variant. For matched filter dependent models, we use models released in \cite{ruzickaOperational} using the U-Net architecture.

\subsection{On-board deployment}

During the Tokyo-FC campaign, NASA's Langley Research Center Gulfstream-III (N520NA) used the AVIRIS-5 instrument. In addition to data collection, on-board preview products described in \cite{thompson2015real_time_mf_on_plane} were also computed. Specifically, preview MF products were computed for CH$_{4}$, CO2 and NO$_{2}$. We were able to extend the preview computing resource for additional post-processing with our ML models. 
The computer on-board has the 6-core Intel(R) Xeon(R) Gold 6342 CPU @ 2.8GHz, a NVIDIA RTX A4000 with 16GB GPU RAM and has around 128GB free RAM. As such we highlight that it is a much more capable compute unit than our testbed device. 

Each 67.5 seconds, 10,000 raw lines of 1239 samples are collected during the flight. Lines are binned 5 times resulting in the final data shape of one scene of 1239 samples x 2000 lines (each at all spectral bands). MF product is computed for each explored trace gas, and for methane the single band product is used with our ML model. Our data ingestion script pads the input shape to a multiple of 32, resulting in 1248x2016 px per scene.

\section{Results}

\subsection{Testbed timing}

On our testbed device, we compare the full version of L1A to L1B data processing with a variant which skips selected steps.
For easier comparison with the deployment platform, we consider the same spatial resolution in both cases per scene.
While the full on-the-ground processing pipeline (only skipping OSF and keeping all other steps) takes about 2088.45 seconds, our sped up variant written in Numpy takes only 2.61 seconds on CPU and the PyTorch GPU variant 0.46 seconds.
Matched filter computation adds between 76.78-122.85 seconds per whole scene on our testbed.


Finally, the machine learning model inference is estimated at around 1.84 seconds per scene when using only the matched filter product as input. This increases to about 2.15 seconds when the L1B data is used instead - however we highlight that this avoids the computational bottleneck in the form of the matched filter product computation.




\subsection{On-board detection of methane point source emission}

From the flight logs recorded during the Tokyo-FC campaign, we report the ML inference times. It takes approximately 2.59$\pm$0.93 seconds to make a prediction for an entire scene. 
This is slower than on our lower compute testbed device, which is likely due to the other processes running on the device at the same time during a real mission.
This is approximately 67.5 seconds worth of data, which means that we are safely under the requirements of the mission. Importantly, model inference adds only minimal time to other already executed processing steps.
Figure \ref{fig:detection} details the correctly detected methane point source event observed during the Tokyo-FC campaign on the 11th March 2026 when flying over an active power plant in Kawasaki, Japan.

\section{Conclusion}

We present initial results from the deployment of machine learning models on-board of aerial platform used in the Tokyo-FC campaign. We present what we believe to be the first automated detection of methane leak on-board with machine learning, a significant and logical next step to the work of \cite{thompson2015real_time_mf_on_plane}. Predictions of our model, alongside with the collected data are available in the public portal at\footnote{\url{https://popo.jpl.nasa.gov/mmgis-aviris/?mission=TokyoFC}}.
As future research, we aim to explore the usage of machine learning models recently proposed in \cite{ruzicka2026trace_gases} for multi-species detection of trace gases.

\bibliographystyle{IEEEtran}
\bibliography{biblio}

\end{document}